\documentclass[11pt,a4paper]{article}
\usepackage[hyperref]{naaclhlt2018}
\usepackage{times}
\usepackage{latexsym}
\usepackage{booktabs}
\usepackage{url}
\usepackage{graphicx}
\usepackage{cuted}
\usepackage{caption}
\usepackage{listings}
\usepackage{float}

\aclfinalcopy 

\title{Bidirectional Recurrent Models for Offensive Tweet Classification\\ 
}

\author{Aleix Cambray Roma \\
  Imperial College London\\
  Department of Computing\\
  {\tt ac7914@ic.ac.uk} \\\And
  Norbert Podsadowski\\
  Imperial College London\\
  Department of Computing\\
  {\tt np1815@ic.ac.uk}}

\date{}

\begin{document}
\maketitle

\begin{abstract}
  In this paper we propose four deep recurrent architectures to tackle the task of offensive tweet detection as well as further classification into targeting and subject of said targeting. Our architectures are based on LSTMs and GRUs, we present a simple bidirectional LSTM as a baseline system and then further increase the complexity of the models by adding convolutional layers and implementing a split-process-merge architecture with LSTM and GRU as processors. Multiple pre-processing techniques were also investigated. The validation F1-score results from each model are presented for the three subtasks as well as the final F1-score performance on the private competition test set. It was found that model complexity did not necessarily yield better results. Our best-performing model was also the simplest, a bidirectional LSTM; closely followed by a two-branch bidirectional LSTM and GRU architecture.
\end{abstract}

\section{Introduction}

The main task of OffensEval-2019 \cite{offenseval} is to detect and classify offensive language in social media, specifically tweets. This task is partitioned into three subtasks, involving the classification of anonymised tweets.
\begin{itemize}
    \item Task A: Offensive/Not Offensive
    \item Task B: Of those that are offensive, whether they are targeted or not targeted.
    \item Task C: Of those that are targeted, whether they are targeted at an individual, at an organisation or other.
\end{itemize}

For these, several recurrent neural network architectures were implemented, as presented in Section 3. The models were subsequently optimised and their performances compared in order to converge to the best model for each task. The data was also pre-processed before being fed to the model as discussed in Section 2. The results are given in Section 4 and discussion of these as well as of challenges encountered is presented in Section 5.

\section{Data}

\subsection{Data Handling and Preprocessing}

Training data for the competition \cite{OLID} was given in a plaintext format of tab-seperated values consisting of \textit{tweet ID} and \textit{tweet content}, labelled with 3 values for \textit{task A} = $\{OFF, NOT\}$, \textit{task B} = $\{TIN, UNT\}$ and \textit{task C} = $\{IND, GRP, OTH\}$.

Prior to training the models, we preprocessed the data extensively. The primary aim of this is to ensure that models were trained on the most normalized representation of tweets possible, allowing us to take full advantage of our pretrained word embeddings, with the overall goal of increasing model performance as much as possible.

This section outlines all the data handling and processing techniques carried out.

\subsubsection{User mentions and URLs}

\textit{Tweet content} data consisted of anonymised Twitter user mentions in the form of \verb|@USER|. For example, the trial data (without anonmyzied mentions) contained this row: \verb|Hey @LIRR , you are disgusting|. One could add a module to the classifier that initially looks up @LIRR on Twitter, learns that it is the Long Island Rail Road, and helps with the classification that this tweet was \textit{targeted} at an \textit{organization} and \textit{offensive}. Unfortunately, the training data anonymised all of these user mentions which intuitively would be of little value to the model, and thus all such mentions were removed.

Similarly, all instances of real URLs were removed and changed into \verb|URL|. One could argue that a more complicated classifier could make use of such information, but in its anonymized form we have decided to remove these as well.

\subsubsection{HTML entities}
The dataset contained certain HTML entities such as \verb|&gt;| that represent certain special characters. These do not contribute to the meaning of the tweets, and thus were removed.

\subsubsection{Hashtags}

Even though hashtags can contain relevant information, their verbal form is complex to deal with. We decided to simply remove them - with the amount of data we have, it is unlikely that we have a lot of tweets where the hashtag influences the meaning of the tweet.

\subsubsection{Lowercasing}

 One could argue that uppercase words could have a more offensive meaning than its lowercase version. Given this small dataset, this would not occur very frequently; we have therefore decided to reduce the noise by normalising each word to be lowercase, rather than introducing extra noise for the model to deal with.

\subsubsection{Non-ASCII filtering}

The next step was to delete all non-ASCII characters. Tweets could contain all kinds of non-ASCII characters: primarily emojis and non-Latin characters. For the former, we decided to not consider these in the model, following similar reasoning as before - emojis would not occur often enough to warrant a more sophisticated approach. Non-Latin characters were also found very rarely, as the data set given was intended to contain only English tweets, hence these characters can be discarded as noise.

Further, we wanted to avoid having the same word once alone and once followed by a Non-Latin character, which would not be considered as the same word, and tagged as unknown. To alleviate this issue, we partitioned and added spaces around non-Latin characters, such as symbols. For instance, \verb|me+you=forever| would be transformed into \verb|me + you = forever|.

\subsubsection{Apostrophe handling}

Most of the the tweets contain contract verbal forms. This creates noise and hides negation which is important for offensive detection. For instance we transformed \verb|aren't| into  \verb|are not| or \verb|i'm| into \verb|i am|. Then the verbs were also fed in the lemmatiser, detailed later on.
\subsubsection{Punctuation removal}
We removed punctuation such as question marks, commas, colons and periods. 
\subsubsection{Number removal}
Numbers are irrelevant for this problem so we decided to remove them completely. For instance \verb|Obama2020| is transformed into \verb|Obama|.

\subsubsection{Stop word removal}

We removed all stop words such as \textit{is, that, the} etc. as they contribute little meaning to the tweets, The NLTK stop-word set\footnote{https://gist.github.com/sebleier/554280} was used.

\subsubsection{Reduction of word lengths}

Following tokenisation, we used a technique known was word length reduction based on the fact that all English words allow a maximum of two consecutive character repetitions. For example, we correct words such as \verb|realllllly| to \verb|really| or \verb|aaaaaaaaaaaah| to \verb|aah|. This allows us to normalize such occurrences of words prior to the following dictionary-based pre-processing techniques.

\subsubsection{Word segmentation and spelling correction}

We perform word segmentation using a fast state-of-the-art library SymSpell\footnote{https://github.com/mammothb/symspellpy}. This uses a Triangular Matrix approach to correct words such as \verb|thecatonthemat| to \verb|the cat on the mat| with extremely good accuracy and high performance. 

We then use the same SymSpell library and its Symmetric Delete approach to very quickly correct word spellings. We pick the closest matching word within an edit distance of 2.

As a pre-requisite to both the word segmentation and spelling correction, we use our corpus of words to create a frequency-aware dictionary that SymSpell uses to guide its segmentation and spell-checking processes.

\subsubsection{Lemmatisation}

Finally, we lemmatise all words to their base representation, such \verb|killing| to \verb|kill| or \verb|eating| to \verb|eat|. This is important in ensuring that the meaning of words is maintained regardless of their exact form displayed in the sentence, increasing model performance. The NLTK package \cite{nltk} was used for this step.

\subsubsection{Sentence padding}

A requirement for our LSTM models was that the input sentence length has to be fixed for all training examples. We therefore had to pick an appropriate sentence padding/truncation length $p$. All sentences with more than $p$ words would be truncated to $p$ and all sentences with less than $p$ words would be padded with a reserved padding token.

\begin{figure}[H]
  \includegraphics[width=\linewidth]{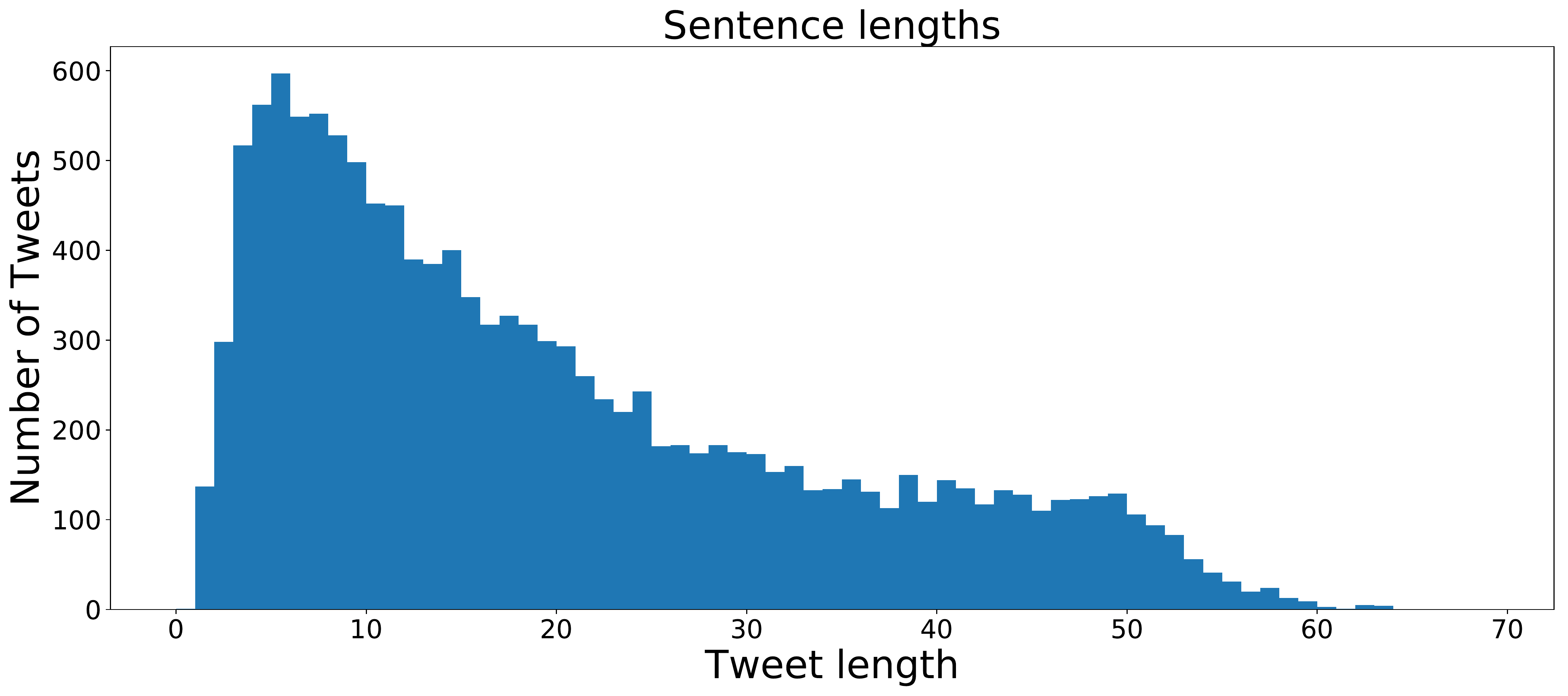}
  \caption{Distribution of sentence lengths after pre-processing all tweets.}
  \label{fig:lengthhistogram}
\end{figure}

After the aforementioned data processing tasks, we plotted the distribution of tweet lengths on a histogram in order to guide our decision on $p$. This showed that a majority of the sentences had sentence lengths in the 3-50 words range, as shown in Figure \ref{fig:lengthhistogram}.

The decision on $p$ represents a trade-off: a low value of $p$ means that we lose a lot of words in the longer sentences, but the shorter sentences need less padding. It was unclear whether amount of padding in shorter sentences would have a significant effect on the models; however, we knew that truncating a lot of long sentences would lose a lot of information. We took a conservative approach by setting $p$ to 50, which captured most of our sentence lengths.
 
\subsection{Data Representation: Word-embeddings}

We tried two approaches for our data representation.

\begin{itemize}
    \item Learnt word embeddings:\\
    We trained our own word embeddings during the execution of the whole model, with a randomly initialised embedding layer.
    \item Pre-trained GloVe Embeddings \cite{pennington2014glove}:\\
    We used GloVe Twitter 27B embeddings\footnote{http://nlp.stanford.edu/data/glove.twitter.27B.zip}. These were trained from a corpus of 2 billion tweets, with 27 billion tokens. The vocabulary size is 1.2 million words, and the embedding dimensionality was 100.
\end{itemize}

All of our models seemed to perform much better using the pre-trained GloVe word embeddings. One explanation for this is that such embeddings are much richer in content and embed inter-word semantical correlations much better than what could be learnt during learning the classifiers using the limited provided dataset. 

\section{Model Design and Training}

We devised four different deep recurrent network architectures, which we tested on all three tasks. \textit{Keras} \cite{chollet2015keras} was used as the high-level library to implement these, with a TensorFlow \cite{tensorflow2015-whitepaper} backend.

The simplest model we trained was a recurrent bidirectional LSTM model. LSTM \cite{hochreiter1997long} was chosen over RNN in order to alleviate the vanishing gradient problem and so the network is able to learn long-term dependencies between different words. A similar reasoning is used to justify the need for a Bidirectionality \cite{schuster1997bidirectional}, in so that there is no algorithmic bias towards later regions of the tweet. The other three architectures build upon this one are of increasing complexity.

\subsection{biLSTM}
\begin{figure}[H]
  \includegraphics[width=\linewidth]{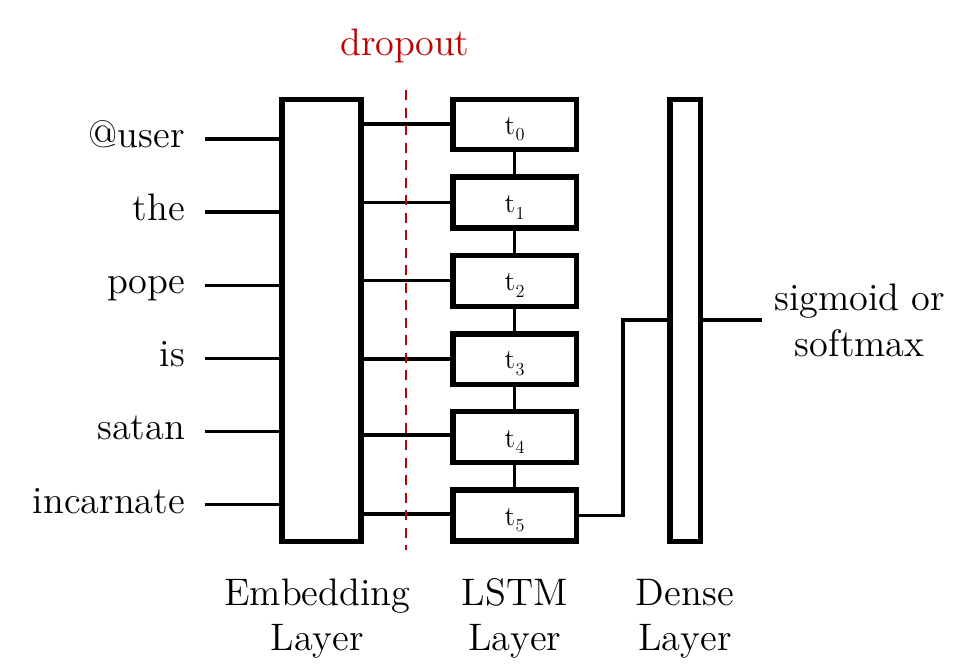}
  \caption{Schematic of the bi-LSTM Architecture.}
  \label{fig:bilstm}
\end{figure}

This model, depicted in Figure \ref{fig:bilstm},  as well as the other three, takes as input the word embeddings as a matrix which is comprised of the vertically stacked embedding vectors corresponding to the words present in the tweet. This matrix can be thought of a sequence of embedded words. Each of these embedding vectors is fed to the bidirectional LSTM at their respective timestep and the final timestep output is then connected to a dense layer with sigmoidal or softmax-activated neurons depending on the Task.
One-dimensional Spatial dropout was added between the embedding layer and the LSTM. This type of dropout scheme drops entire rows of the embedded matrix, equivalent to dropping words from the sequence. The rationale behind this is to discourage the network to rely on specific words from the training set and therefore introduce more language-specific regularisation. This is in contrast to normal dropout which would randomly drop connections from specific elements in the embedded matrix which has no grammatical interpretation.

\subsection{CNN-biLSTM}
\begin{figure}[H]
  \includegraphics[width=\linewidth]{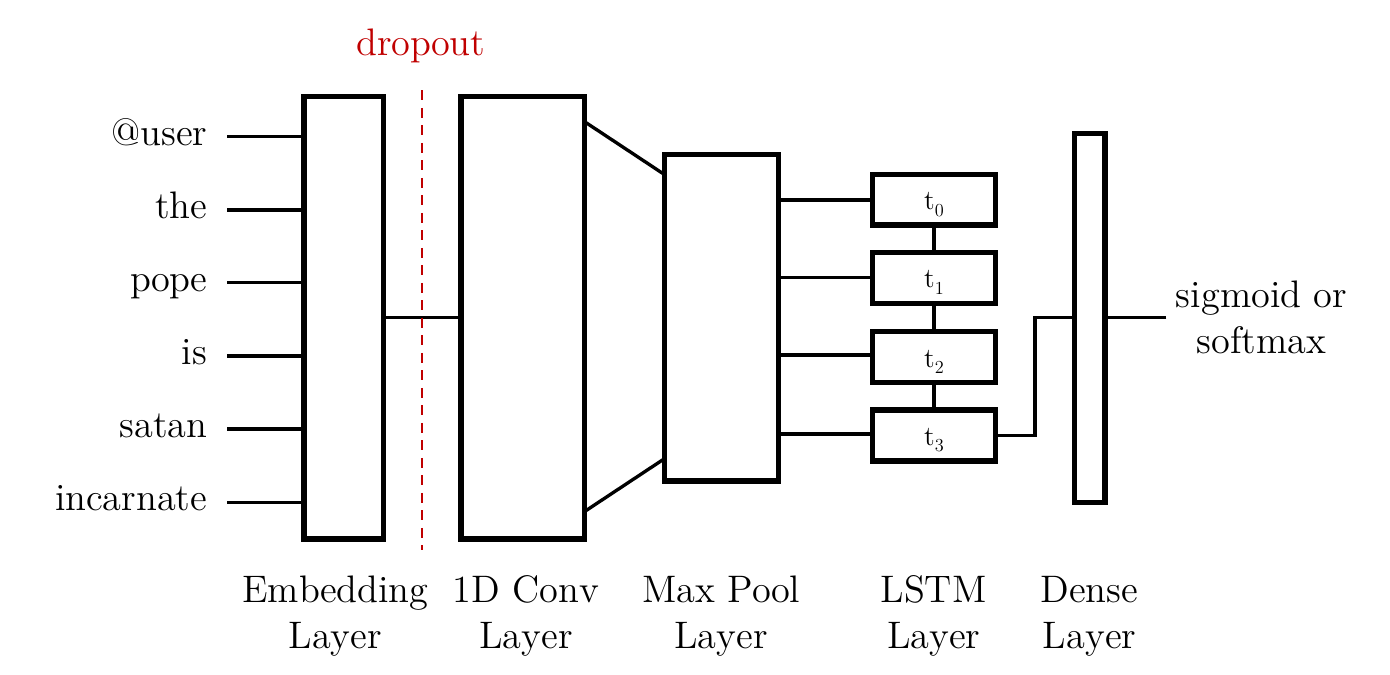}
  \caption{Schematic of the CNN-biLSTM Architecture.}
  \label{fig:cnn-bilstm}
\end{figure}

The CNN-biLSTM is depicted in Figure \ref{fig:cnn-bilstm}. In this case, a 1D convolutional layer was used to convolve the embedding vectors along the temporal dimension with a kernel of size 4 and 64 output filters. This layer uses ReLU activation and was therefore initialised with a He uniform weight distribution. This convolutional layer is followed by pooling with pool size 4 and stride 4, which decreases dimensionality before feeding the resultant sequences to the Bidirectional LSTM layer. Again, the output vector is then fed to a dense layer with 1 sigmoidal output neuron (Tasks A and B) or 3 softmaxed output neurons (Task C).
Adding this convolutional layer first means the inputs to the LSTM are local combinations of words, rather than individual words. While this adds complexity to the network, it is not a linguistically motivated change. CNN-LSTMs have been found to be useful for tasks with spatial inputs such as image sequences or 3-D point cloud sequences, in which local features are very relevant.

\subsection{biLSTM-CNN}
\begin{figure}[H]
  \includegraphics[width=\linewidth]{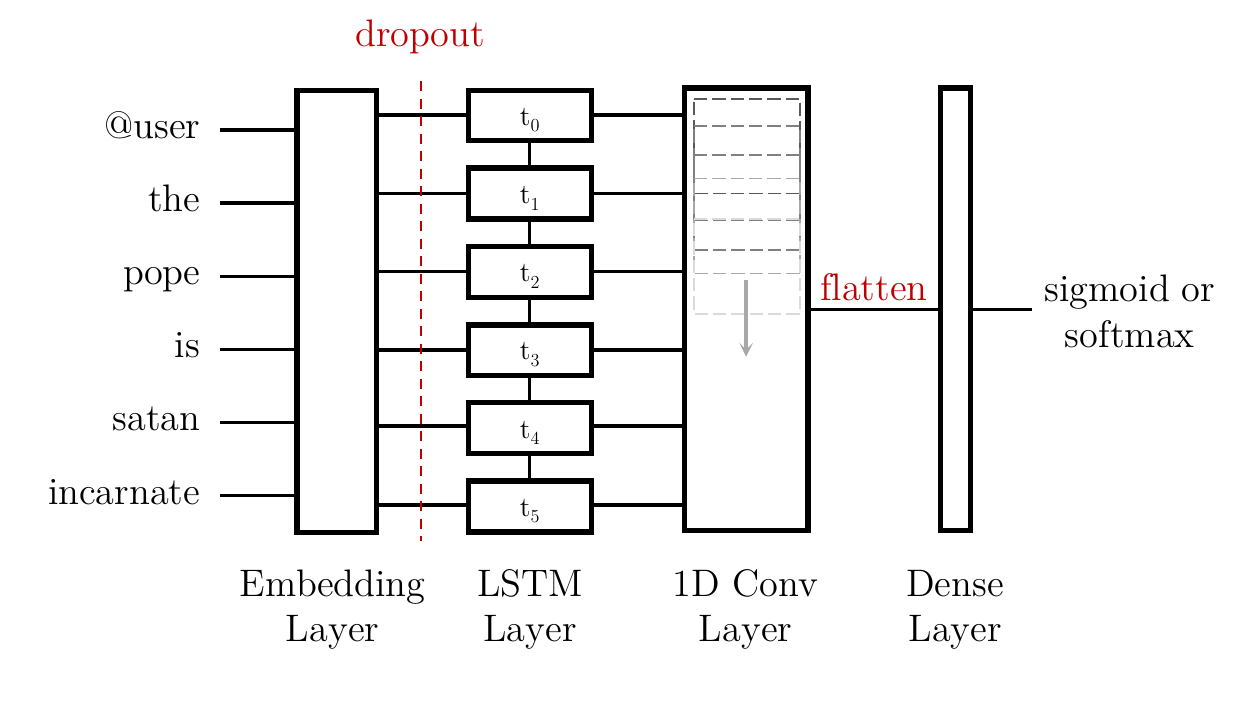}
  \caption{Schematic of the biLSTM-CNN Architecture.}
  \label{fig:bilstm-cnn}
\end{figure}

The biLSTM-CNN, depicted in Figure \ref{fig:bilstm-cnn} has the convolutional layer moved after the bidirectional LSTM. For this architecture, the LSTM was set to output its state for each timestep in order for the convolutional layer to be able to convolve along a temporal axis. This is in contrast to the previous two architectures, in which only the latest timestep output was used. No pooling was used after convolution, so to not unnecessarily lose meaningful information. The reasoning behind convolving after the LSTM is to further combine the long-term dependencies the LSTM has uncovered and add expressivity to the network. Therefore it is expected that this will perform better than the limiting CNN-biLSTM architecture.

\subsection{biGRU $\oplus$ biLSTM}

\begin{figure}[H]
  \includegraphics[width=\linewidth]{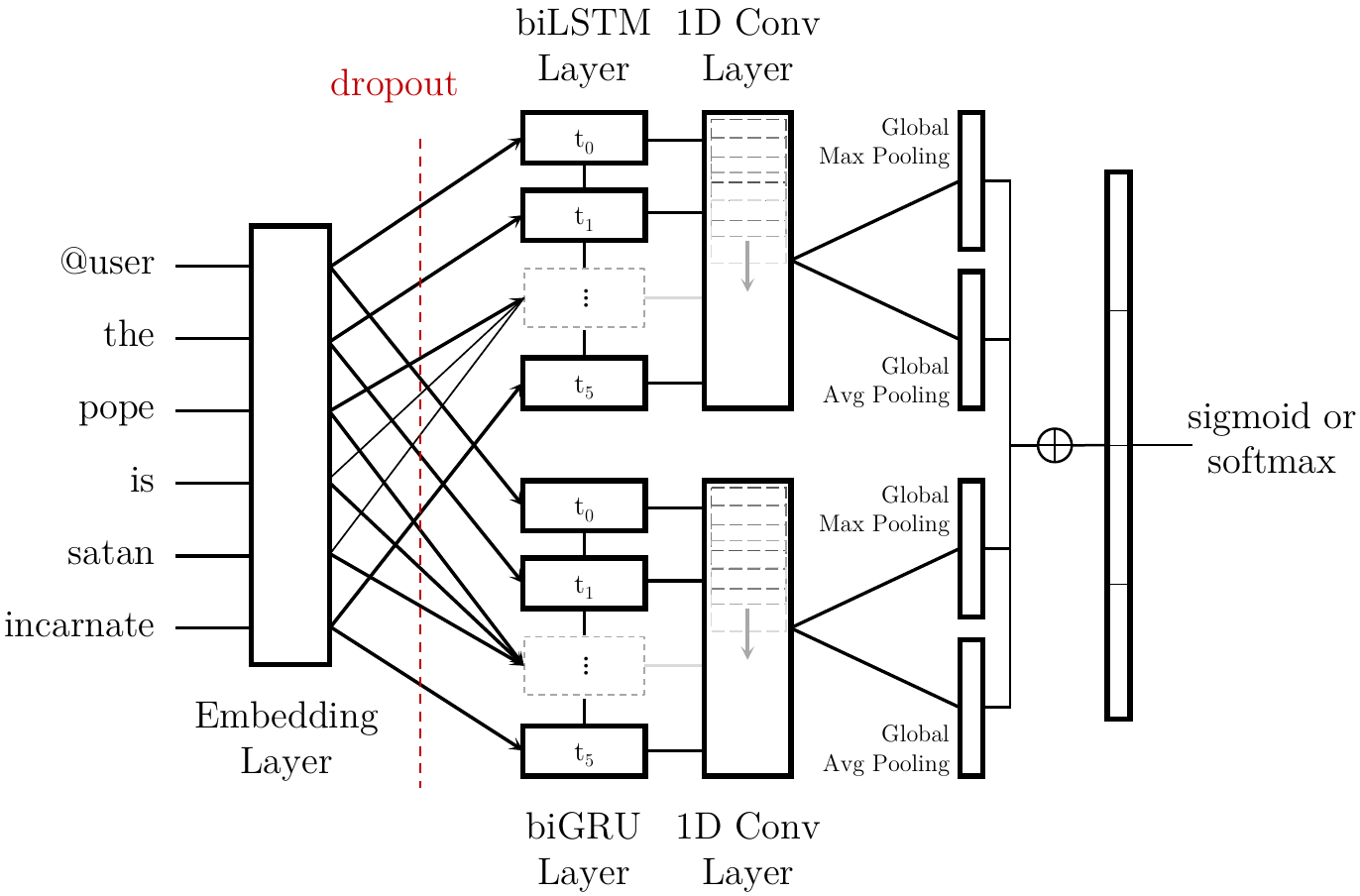}
  \caption{Schematic of the bi GRU LSTM Architecture.}
  \label{fig:gru_lstm}
\end{figure}

Finally, introduced gated recurrent units (GRUs) \cite{gru}, as they have been shown to, in some cases, improve performance relative to LSTMs as well as being computationally more efficient \cite{grushow}. Since it is inconclusive in the literature which is to perform best, we devised an architecture which combines both. This is depicted in Figure \ref{fig:gru_lstm}. The embedded words are processed in parallel through two branches of biLSTM-CNN and biGRU-CNN. Global max pooling and global average pooling is then applied to each CNN output and the four resulting vectors are concatenated to form a 1D array which is then fed to a dense layer with 1 sigmoidal unit or 3 softmaxed units depending on the task. This architecture adds more expressivity to the network as well as allowing it to squeeze the best from both types of recurrent cells.

\section{Training procedure}
\subsection{Addressing class imbalance}

For all tasks, our training dataset was very unbalanced: e.g. for Task B, 88\% of all labels were the $TIN$ class. We tackled this using \textit{class weighting}. The loss function was weighted by the inverse of the proportion of each class in the training set; this means that weight updates during any single pass in our networks would have a greater effect when training under-represented classes. This technique worked well to improve overall performance by reducing overfitting.

Prior to this, we considered other approaches. \textit{Down-sampling} is a technique which would discard examples from over-represented classes. Here, the gain in balancing the data-set did not outweigh the loss in having a very small number of examples to train on (due to the small data set), and thus resulted in very poor performance. 

We also thought about using \textit{naive over-sampling}, which would replicate examples in under-represented classes - intuitively, this likely would have had a similar effect to class weighting, but would be harder to implement and performant.

A more interesting technique is \textit{over-sampling using Smote}, which involves generating synthetic training examples from real-valued vectors. We could use the embedding matrices for under-represented sentences in order to generate similar, new vectors not present in our dataset labelled under the same class. Generating synthetic data in this way has potential to allow the model to generalise better and can be investigated in further work.

\subsection{Validation set}

We used an 80\%-20\% split of our dataset into the training and validation sets. We used stratification during this process, which ensures that both sets have a roughly equal proportion of every class, to allow for more representative validation.

\subsection{Optimisation}

We used the Adam optimizer, which improves on the RMS-prop and AdaGrad back-propagation algorithms. We  used the binary cross-entropy loss function for tasks A and B, as well as categorical cross-entropy for task C, since it is a multi-class classification problem.

We also used an early stopping strategy based on a long-term moving-average of the F1 score evaluated at the end of every epoch.

\subsection{Hyperparameter optimization}

Our set of hyperparameters consisted of the following: \\
- Learning rate  \\
- Learning rate decay  \\
- Number of epochs \\
- Batch size \\
- Dropout rate \\
- Recurrent units  \\

We performed hyperparameter optimization using manual tweaking over successive runs on the validation set.. We found that the following parameters yielded the best validation performance:

\begin{verbatim}
LR               = 0.001
LR_DECAY         = 0.001
EPOCHS           = 50
BATCH_SIZE       = 32
DROPOUT          = 0.5
BIDIRECTIONAL    = True
RECURRENT_UNITS  = 50
\end{verbatim}

Given more compute power and time, we would have preferred to run an extensive grid search over the entire hyperparameter space.

\section{Results and Discussion}

\begin{table*}
\centering
\caption{Holdout-validation macro F1 scores and accuracy of all models for the three different tasks.}
\begin{tabular}{@{}lllllll@{}}
\toprule
                               & \multicolumn{2}{c}{\textbf{Task A}}  & \multicolumn{2}{c}{\textbf{Task B}}   & \multicolumn{2}{c}{\textbf{Task C}}     \\ \midrule
                               & \textbf{F1}      & \textbf{Acc}      & \textbf{F1}      & \textbf{Acc}      & \textbf{F1}      & \textbf{Acc}          \\ \cmidrule(l){2-7} 
\textbf{biLSTM}                & 0.7382           & 0.7801            & 0.6115              & 0.5753                  & 0.5001              & 0.6966                      \\
\textbf{CNN-biLSTM}            & 0.6346           & 0.6473            & 0.5521                &  0.5068               & 0.4322                & 0.6142                      \\
\textbf{biLSTM-CNN}            & 0.7170           & 0.7562            & 0.5496               & 0.4903                 & 0.4738                & 0.6863                      \\
\textbf{biGRU $\oplus$ biLSTM} & 0.7285           & 0.7544            & 0.5963               & 0.5722                 & 0.5052                & 0.7012                      \\ \bottomrule
\end{tabular}
\label{table:valresults}
\end{table*}

We gained a few key insights from our results, which are shown in Table \ref{table:valresults}. For most tasks, the most simple biLSTM model outperformed the other architectures, with the more complex biGRU$\oplus$biLSTM closely following. We learned that, at least for this task and data set, \textbf{more complex models did not necessarily result in better performance}. The biLSTM model's macro F1 scores on the OffensEval private test set were 0.77 for Task A, 0.64 for Task B and 0.52 for Task C, In hindsight, however, the model submitted was not optimal as it was not re-trained on the entire dataset and therefore only used 80\% of the training data. We speculate that re-training the model on 100\% of the provided dataset would have yielded significantly better results.

The key to any machine learning model is quantity and quality of data. Perhaps this ~18,000 sample dataset has an insufficient amount of data for the networks to be able to generalise to the wider population of tweets, especially more the complex ones. The dataset was also highly unbalanced, which is one factor we can use to explain the lower performance attained in Task B and C when compared to Task A.

\begin{figure}[]
  \includegraphics[width=\linewidth]{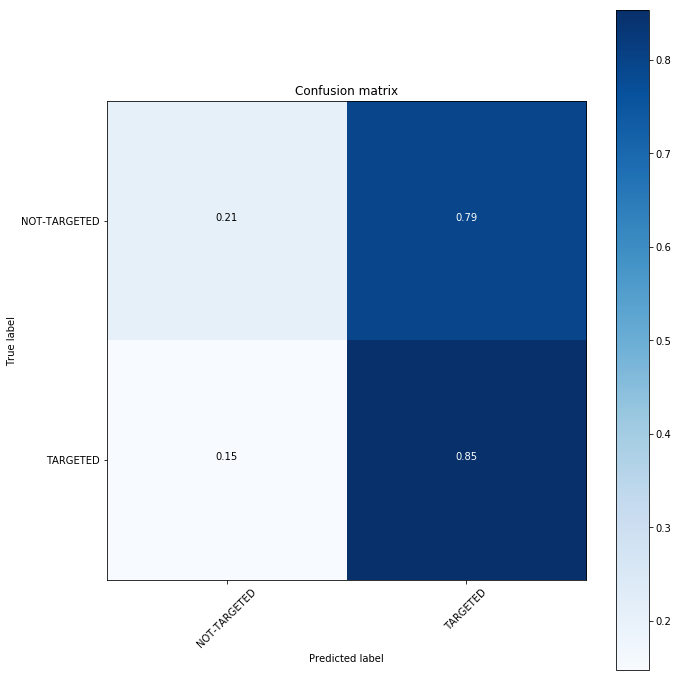}
  \caption{Problematic confusion matrix for Task B, over-prediction of TARGETED due to class imbalance. This problem was alleviated by class weighting.}
  \label{fig:confusion_matrix}
\end{figure}

\begin{figure}[]
  \includegraphics[width=\linewidth]{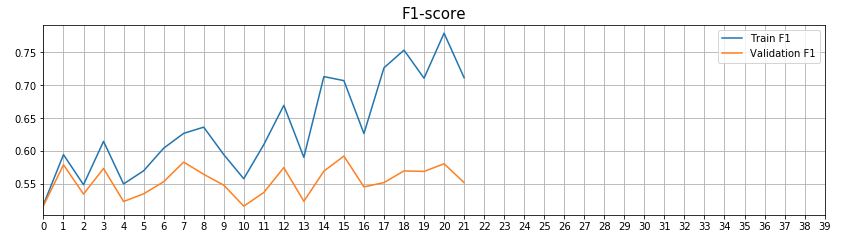}
  \caption{Example of F1 score profile during training for Task B. Blue: Training; Orange: Validation}
  \label{fig:f1_score}
\end{figure}

During training, we monitored how F1 score changed between epochs for both the training set and the validation set. We observed that the F1 score oscillated for a majority of epochs and often stabilized during later epochs (see Figure \ref{fig:f1_score}). One explanation is that while learning on successive mini-batches, the model gets pushed in different directions, but trends towards better performance with the progression of epochs. The F1-score also seldom correlated with the accuracy or loss. One explanation for this lack of correlation is that the F1-score into account the inherent imbalance in our classes by averaging per-class metrics, but the accuracy metric is the product of our loss function minimization which is \textbf{more vulnerable} to the imbalance.
This imbalance originally resulted in models which only or over-predicted the class which was over-represented in the training set, which is a common failure resulting from having cross-entropy as a loss function and getting stuck in a local optimum, an example confusion matrix of this happening is shown in \ref{fig:confusion_matrix}. This was alleviated by weighting the cross-entropy loss function by the inverse of the class support in the training dataset, therefore penalising incorrect predictions of the under-represented classes more harshly and avoiding an "all-on-red" situation.

Given the instability of the F1 score, settling on an architecture and a set of optimal hyperparameters was very tedious and challenging. Given more computational resources and time, we would have liked to run a substantial grid search across a sufficiently extensive hyperparameter space as a more reliable way of picking the best model.

\section{Future work}
Several paths for further work are possible, here are some which are interesting and could yield significant performance improvements.

\begin{itemize}
    \item Gather more data or utilize other datasets
    \item Investigate learning rate scheduling, decay and decrease-on-plateau
    \item Further invistigate Over-sampling both at language-level and at vector-level with SMOTE
    \item Automated sequential hyperparameter optmisation using bayesian optimisation. Or fine-grain grid search if computationally feasible. Since the dataset is not extensive, this approach could potentially be nested within k-fold cross-validation to get a good estimate of the generalisation error.
    \item Work at character-level (robust against misspellings and obscure words)
    \begin{itemize}
        \item Simple: char-CNN or char-LSTM.
        \item Better: char-CNN as word encoder then feed each char-encoded word to a Bidirectional LSTM, use last LSTM output to classify.
    \end{itemize}
    \item Ensembling of different models. An ensemble of very different architectures could prove beneficial. The weak learners could be word-level biLSTMs, character-level LSTMs, more complex models such as BERT or ELMO, traditional approaches such as SVMs, etc. 
\end{itemize}

\bibliographystyle{acl_natbib}
\bibliography{SemEval19_Cambray_Podsadowski}

\begin{thebibliography}{10}
\expandafter\ifx\csname natexlab\endcsname\relax\def\natexlab#1{#1}\fi

\bibitem[{Abadi et~al.(2015)Abadi, Agarwal, Barham, Brevdo, Chen, Citro,
  Corrado, Davis, Dean, Devin, Ghemawat, Goodfellow, Harp, Irving, Isard, Jia,
  Jozefowicz, Kaiser, Kudlur, Levenberg, Man\'{e}, Monga, Moore, Murray, Olah,
  Schuster, Shlens, Steiner, Sutskever, Talwar, Tucker, Vanhoucke, Vasudevan,
  Vi\'{e}gas, Vinyals, Warden, Wattenberg, Wicke, Yu, and
  Zheng}]{tensorflow2015-whitepaper}
Mart\'{\i}n Abadi, Ashish Agarwal, Paul Barham, Eugene Brevdo, Zhifeng Chen,
  Craig Citro, Greg~S. Corrado, Andy Davis, Jeffrey Dean, Matthieu Devin,
  Sanjay Ghemawat, Ian Goodfellow, Andrew Harp, Geoffrey Irving, Michael Isard,
  Yangqing Jia, Rafal Jozefowicz, Lukasz Kaiser, Manjunath Kudlur, Josh
  Levenberg, Dan Man\'{e}, Rajat Monga, Sherry Moore, Derek Murray, Chris Olah,
  Mike Schuster, Jonathon Shlens, Benoit Steiner, Ilya Sutskever, Kunal Talwar,
  Paul Tucker, Vincent Vanhoucke, Vijay Vasudevan, Fernanda Vi\'{e}gas, Oriol
  Vinyals, Pete Warden, Martin Wattenberg, Martin Wicke, Yuan Yu, and Xiaoqiang
  Zheng. 2015.
\newblock \href {http://tensorflow.org/} {{TensorFlow}: Large-scale machine
  learning on heterogeneous systems}.
\newblock Software available from tensorflow.org.

\bibitem[{Cho et~al.(2014)Cho, van Merrienboer, G{\"{u}}l{\c{c}}ehre, Bougares,
  Schwenk, and Bengio}]{gru}
Kyunghyun Cho, Bart van Merrienboer, {\c{C}}aglar G{\"{u}}l{\c{c}}ehre, Fethi
  Bougares, Holger Schwenk, and Yoshua Bengio. 2014.
\newblock \href {http://arxiv.org/abs/1406.1078} {Learning phrase
  representations using {RNN} encoder-decoder for statistical machine
  translation}.
\newblock \emph{CoRR}, abs/1406.1078.

\bibitem[{Chollet et~al.(2015)}]{chollet2015keras}
Fran\c{c}ois Chollet et~al. 2015.
\newblock Keras.
\newblock \url{https://keras.io}.

\bibitem[{Chung et~al.(2014)Chung, G{\"{u}}l{\c{c}}ehre, Cho, and
  Bengio}]{grushow}
Junyoung Chung, {\c{C}}aglar G{\"{u}}l{\c{c}}ehre, KyungHyun Cho, and Yoshua
  Bengio. 2014.
\newblock \href {http://arxiv.org/abs/1412.3555} {Empirical evaluation of gated
  recurrent neural networks on sequence modeling}.
\newblock \emph{CoRR}, abs/1412.3555.

\bibitem[{Hochreiter and Schmidhuber(1997)}]{hochreiter1997long}
Sepp Hochreiter and J{\"u}rgen Schmidhuber. 1997.
\newblock Long short-term memory.
\newblock \emph{Neural computation}, 9(8):1735--1780.

\bibitem[{Loper and Bird(2002)}]{nltk}
Edward Loper and Steven Bird. 2002.
\newblock Nltk: The natural language toolkit.
\newblock In \emph{In Proceedings of the ACL Workshop on Effective Tools and
  Methodologies for Teaching Natural Language Processing and Computational
  Linguistics. Philadelphia: Association for Computational Linguistics}.

\bibitem[{Pennington et~al.(2014)Pennington, Socher, and
  Manning}]{pennington2014glove}
Jeffrey Pennington, Richard Socher, and Christopher~D. Manning. 2014.
\newblock \href {http://www.aclweb.org/anthology/D14-1162} {Glove: Global
  vectors for word representation}.
\newblock In \emph{Empirical Methods in Natural Language Processing (EMNLP)},
  pages 1532--1543.

\bibitem[{Schuster and Paliwal(1997)}]{schuster1997bidirectional}
Mike Schuster and Kuldip~K Paliwal. 1997.
\newblock Bidirectional recurrent neural networks.
\newblock \emph{IEEE Transactions on Signal Processing}, 45(11):2673--2681.

\bibitem[{Zampieri et~al.(2019{\natexlab{a}})Zampieri, Malmasi, Nakov,
  Rosenthal, Farra, and Kumar}]{OLID}
Marcos Zampieri, Shervin Malmasi, Preslav Nakov, Sara Rosenthal, Noura Farra,
  and Ritesh Kumar. 2019{\natexlab{a}}.
\newblock {Predicting the Type and Target of Offensive Posts in Social Media}.
\newblock In \emph{Proceedings of NAACL}.

\bibitem[{Zampieri et~al.(2019{\natexlab{b}})Zampieri, Malmasi, Nakov,
  Rosenthal, Farra, and Kumar}]{offenseval}
Marcos Zampieri, Shervin Malmasi, Preslav Nakov, Sara Rosenthal, Noura Farra,
  and Ritesh Kumar. 2019{\natexlab{b}}.
\newblock {SemEval-2019 Task 6: Identifying and Categorizing Offensive Language
  in Social Media (OffensEval)}.
\newblock In \emph{Proceedings of The 13th International Workshop on Semantic
  Evaluation (SemEval)}.

\end{thebibliography}

\end{document}